\documentclass[letterpaper, 10pt, journal, twoside]{IEEEtran}      % Use this line for a4
\IEEEoverridecommandlockouts                              % This command is only
\usepackage{cite}
\usepackage{amsmath,amssymb,amsfonts}
\usepackage{algorithmic}
\usepackage[ruled,vlined,linesnumbered]{algorithm2e}
\usepackage{graphicx}
\usepackage{textcomp}
\usepackage{xcolor}
\usepackage{nicefrac}
\usepackage{comment}
\usepackage{wrapfig}
\usepackage{multirow}
\usepackage{makecell}
\usepackage[outercaption]{sidecap}
\usepackage{caption}
\usepackage{bm,times}
\usepackage{marginnote}
\usepackage{autonum}
\usepackage{blindtext}
\usepackage[detect-weight=true, detect-family=true]{siunitx}
\usepackage{url}
\usepackage{pifont}
\def\BibTeX{{\rm B\kern-.05em{\sc i\kern-.025em b}\kern-.08em
    T\kern-.1667em\lower.7ex\hbox{E}\kern-.125emX}}

\let\oldtwocolumn\twocolumn
\renewcommand\twocolumn[1][]{%
    \oldtwocolumn[{#1}{
    \begin{center}
        \includegraphics[width = 1\textwidth]{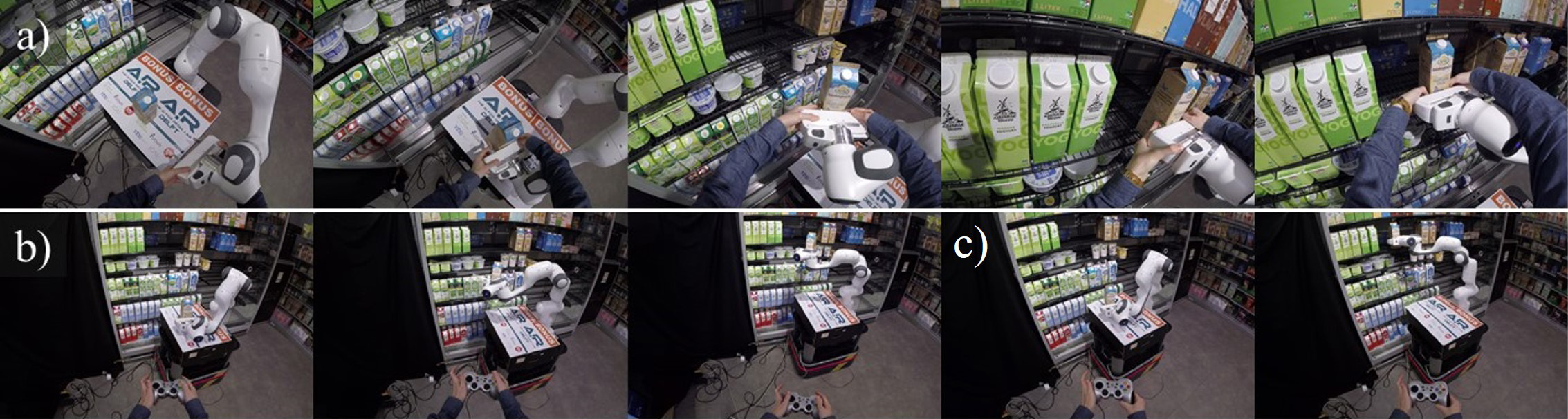}
        \captionof{figure}{Learning flow for teaching a robot how to reshelve an item.; a) starting with a single demonstration, followed by b) multiple rounds of correction after which, c) the robot is able to autonomously carry out the task.}
        \label{fig:learningFlow}
        \end{center}
    }]
}

\begin{document}

\title{Learning to Pick at Non-Zero-Velocity from Interactive Demonstrations}
\author{Anna Mészáros, Giovanni Franzese$^*$, and Jens Kober
\thanks{Manuscript received: October 6, 2021; Revised: February 23, 2022; Accepted: March 17, 2022.}
\thanks{This paper was recommended for publication by Editor Markus Vincze upon evaluation of the Associate Editor and Reviewers’ comments.}
\thanks{This research is partially funded by European Research Council Starting Grant TERI ``Teaching Robots Interactively'', project reference 804907. This research was supported by Ahold Delhaize. All content represents the opinion of the authors, which is not necessarily shared or endorsed by their respective employers and/or sponsors.}
\thanks{Authors are with Cognitive Robotics, Delft University of Technology, Mekelweg 2, 2628 CD Delft, The Netherlands (e-mail:\{a.meszaros, g.franzese, j.kober\}@tudelft.nl).}
\thanks{$^*$Corresponding Author}
\thanks{Digital Object Identifier (DOI): see top of this page.}
}

\maketitle
\begin{abstract}
This work investigates how the intricate task of a continuous pick \& place (P\&P) motion may be learned from humans based on demonstrations and corrections. Due to the complexity of the task, these demonstrations are often slow and even slightly flawed, particularly at moments when multiple aspects (i.e., end-effector movement, orientation, and gripper width) have to be demonstrated at once. Rather than training a person to give better demonstrations, non-expert users are provided with the ability to interactively modify the dynamics of their initial demonstration through teleoperated corrective feedback. This in turn allows them to teach motions outside of their own physical capabilities. In the end, the goal is to obtain a faster but reliable execution of the task.
The presented framework learns the desired movement dynamics based on the current Cartesian position with Gaussian Processes (GPs), resulting in a reactive, time-invariant policy. Using GPs also allows online interactive corrections and active disturbance rejection through epistemic uncertainty minimization.  
The experimental evaluation of the framework is carried out on a Franka-Emika Panda. Tests were performed to determine i) the framework's effectiveness in successfully learning how to quickly pick \& place an object, ii) ease of policy correction to environmental changes (i.e., different object sizes and mass), and iii) the framework’s usability for non-expert users.
\end{abstract}

\section{Introduction}

\IEEEPARstart{M}{ore} often than not, robots employ a pick and place (P\&P) strategy wherein they approach the object, stop and grip it and only then resume moving. 
We as humans, on the other hand, tend to pick things in a single fluent and quick motion.
Of course, robots should also be able to complete a task fairly quickly, which in the case of P\&P introduces a number of challenges, both from a control point of view \cite{van2021robot} as well as a learning point of view \cite{billard2019trends}. 

Learning from Demonstration (LfD) has become a popular approach for allowing non-expert users to teach robots and thus more easily integrate them into the working and daily environment \cite{ravichandar2020recent}. Yet these provided demonstrations are sub-optimal compared to what the robot might be able to achieve, e.g., demos having slower dynamics. 
Concurrently, it is important to consider that often, the execution of a task cannot simply be sped up uniformly. 
For example, when learning a P\&P movement, retaining a high velocity when approaching the object can generate high impact forces which can cause the object to bounce away or topple over, potentially damaging the item in question as well as making it impossible to pick on time. We as people are able to identify such constraints and adapt accordingly, and can transfer this knowledge to the robot through demonstrations.

This work studies the feasibility of robot picking only using time-independent policies learned from human demonstrations and corrections. 
Our previous work \cite{Franzese2021ILoSA} already revealed the effective application of minimum uncertainty GPs for learning variable impedance control in force application tasks like cleaning, plugging, and pushing. 
In none of the previous cases, however, were the dynamics of the end-effector (EE) orientation or gripper learned nor were there critical contact dynamics involved. Teaching more degrees of freedom while asking for fast performance makes the task of non-zero-velocity picking a challenging benchmark for studying the potential of learning from non-expert human teachers. 

The main contributions of this work over the previous are:
    \begin{enumerate}
        \item Proposing a framework for interactively altering the speed and shape of robot motion dynamics in a decoupled manner through teleoperated correction.
        \item A novel minimum uncertainty inference for learning the desired non-linear constraints of EE orientation and gripper width w.r.t.\ the EE position dynamics, while avoiding dangerous extrapolations.
        \item Showing the benefit of uncertainty minimisation for enabling local motion consistency when dealing with critical precision tasks like fast picking, while being compliant in the interaction.
        \item Extending the framework for generalizing to different object positions thanks to the parametrization w.r.t.\ moving reference frames.
    \end{enumerate}

Fig.~\ref{fig:learningFlow} summarizes the three phases of learning in the teaching of a re-shelving operation: the initialization of the policy with kinesthetic demonstration, the shaping of the dynamics with teleoperated corrections, and the final evaluation of the autonomous task execution.  

\section{Background and Related Work}
\label{releted_works}
When executing high-speed manipulation tasks which involve establishing contact with an object, it is important to consider the behaviour around the moment of impact. A reoccurring approach observed in existing works consists of adapting the relative velocity in order to mitigate the effects of the impact \cite{nemec2016speed}. Another strategy, which has been employed to absorb impacts particularly in catching tasks \cite{kim2014catching}, involves utilising a follow-through behaviour which continues to track the predicted path of an object even after interception
%\cite{kim2014catching} 
\cite{salehian2016dynamical}.
Alternatively, one can incorporate compliant behaviour into a provided attractor using impedance control \cite{bogdanovic2020learning}. While it is unable to mitigate the initial impact force irrespective of the set stiffness since the main contribution to this force is the velocity of the impacting objects, it is beneficial for absorbing the post-impact forces \cite{haddadin2009requirements}.

We can conclude that matching the velocity of an object likely achieves the best reduction of impact force, however, such an approach may not be optimal when considering the total time of the trajectory execution. This is especially true for static objects, wherein matching velocities would effectively bring the robot to a stand-still prior to the picking action. A better approach, therefore, is to interactively learn the feasible non-zero contact velocity while ensuring moderate impact forces.

Being able to adapt/correct the learned velocity with ease plays a key role in speeding up the overall execution of the demonstrated trajectory while also considering that the movement dynamics may require different degrees of adaptation at different points of the trajectory; for example slowing down prior to the moment of interception. Different works explore speed adaptation during trajectory execution using different function approximators. One approach involves altering the phase rate of probabilistic movement primitives (ProMPs) \cite{koert2019learning}, whereas others propose a modified version of Dynamical Movement Primitives (DMPs) in which the speed is altered through an additional phase-dependent temporal scaling factor \cite{nemec2018human}, or where the temporal scaling factor is changed through corrections and subsequently translated to changes in the learned dynamic movement \cite{kastritsi2018progressive}. 
The mentioned works modulate the velocity either using optimisation approaches or defined functions, or in the case of \cite{kastritsi2018progressive} where human corrections are used, the corrections are provided in a coupled manner for both the trajectory shape and speed. Our approach instead focuses on combining imitation learning and human interactive feedback \cite{perez2020interactive} to provide corrections to speed and shape in a decoupled manner through teleoperation.

An alternative to phase-dependent methods, like DMPs, can be obtained as the formulation of the motion as a reactive controller according to
\begin{equation}
    \dot{\bm{x}}=f(\bm{x})
    \label{eq:ds}
\end{equation}
where $\bm{x}$ is the robot state and $f$ identifies the transition of the robot state. 
GPs have been used for shaping a motion from human demonstrations through the local modification of a stable field \cite{kronander2015incremental}.
However, none of the other works on learning state-dependent dynamical systems take into account the information of the uncertainty to increase motion consistency, and reduce covariate shift. 
Furthermore, in the context of interactive learning, we introduced the idea of decoupling the corrections of shape and velocity and investigated how this can be beneficial for allowing non-expert users to teach challenging tasks.

\section{Methodology}
The goal of this framework is to enable a user to teach the robot the desired motion through demonstration and teleoperated correction, see Alg.~\ref{algo::FG}. The robot is learning the desired minimum uncertainty dynamical system on the end-effector, formalized in Sec.~\ref{sec::dynamicalSystem} and the dynamics of the gripper orientation and width as a function of the current robot position, formalized in Sec.~\ref{sec::mapping}. The main aim is to show that it is possible to learn a policy and later correct the velocity so as to achieve and surpass the performance of a skilled demonstrator. All of these aspects are modelled with Gaussian Processes, allowing interactive corrections of the dynamics and actions online, see Sec.~\ref{sec::interactive}.

\subsection{Learning a Minimum Uncertainty Dynamical System}
\label{sec::dynamicalSystem}
A non-linear dynamical system can be described by \eqref{eq:ds}.
This type of formulation would fit perfectly in a velocity controller, however, due to the necessity of dealing with impacts --- for which an impedance controller is more suitable \cite{haddadin2009requirements} --- we can rewrite the motion dynamics into its integral form, i.e. we are controlling the desired next point of the motion and not the current desired velocity, based on 
\begin{equation}
    \bm{x}_\mathrm{des} =\bm{x}_t+\int_{t}^{t+\Delta t}{ \bm{\dot{x}} \, dt}=
    \bm{x}_t+(\bm{x}_{t+\Delta t}-\bm{x}_{t})
    =\bm{x}_t+\Delta\bm{x}(\bm{x}_t)
\end{equation}
where $\bm{x}_\mathrm{des}$ is the desired attractor position. Since $\dot{\bm{x}}$ is a function of the current position $\bm{x}$, the integral attractor distance $\bm{\Delta x}$ is going to be a function of the robot position $\bm{x}_t$.
The dynamical system %\eqref{eq:ds} 
can be seen as an external (and slower) control loop where the attractor position is updated as a function of the robot position while the inner (and faster) impedance control loop simulates the dynamics of a critically damped second order dynamical system towards the chosen attractor. 
As an analogy to humans, the slower loop can be seen as the intention update when generating a motion according to the current
perceived 
arm position while the impedance control represents the compliance of the muscles and the joints in the interaction with the environment. 

The desired $\Delta\bm{x}$ is fitted with a Gaussian Process (GP) using the data of a kinesthetic demonstration and user-provided corrections.
A GP is a non-parametric regression method \cite{Rasmussen2005} where the mean and variance of the evaluation point are denoted as
\begin{equation}
\label{eq::GP}
    \bm{\mu} = \bm{k_*}(\bm{\xi},\bm{x})^\top \bm{K}(\bm{\xi},\bm{\xi})^{-1} \bm{y} ,
\end{equation}
\begin{equation}
    \Sigma = k(\bm{x},\bm{x})-\bm{k}_{*}(\bm{\xi},\bm{x})^\top \bm{K(\bm{\xi},\bm{\xi})}^{-1}\bm{k}_{*}(\bm{\xi},\bm{x}),
\end{equation}
where $\bm{x}$ is the evaluation point, $\bm{\xi}$ is the input database and $\bm{y}$ is the output database, and $\bm{\mu}$ and $\Sigma$ are the mean and variance of the regression in the evaluation point.   
The chosen kernel function of the process in this study is the sum of an Automatic Relevance Determination Squared Exponential kernel and a White Noise kernel according to 
\begin{equation}
    k({\bm{x}_i}, {\bm{x}_j}) = \sigma_f^2 \hspace{0.1 cm} e^{\left(-\frac{1}{2}({\bm{x}_i}  - {\bm{x}_j})^T \bm{\Theta}({\bm{x}_i}  - {\bm{x}_j}) \right)} + \sigma_n ^2 \delta_{ij}
    \label{eq:SE}
\end{equation}
where $\delta_{ij}$ is the Kroneker delta, $\Theta$ is a diagonal matrix of the horizontal lengthscales, $\sigma_f$ is the vertical lengthscale, and $\sigma_n$ is the observation noise. These hyper-parameters are the result of the likelihood maximization of sampling $ \bm{y} $ from the fitted Gaussian Process. 
To avoid over/underfitting, we employed a constrained optimization between reasonable bounds for the search of the optimal hyperparameters. 

Finally, something to consider when learning a dynamical system in a reactive formulation is that the next robot position is a function of the learned desired transition but also the external disturbances. This may lead the robot in a position where its policy is not confident anymore, i.e., high epistemic uncertainty. Depending on where this occurs, the robot may not be able to successfully pick up the object or bring it to its goal and execute its motion. 
When we, as humans, execute a motion we try to remain in regions where we are confident about what we have learned up to that point. To encode this behaviour also in the robot, the dynamical system was superposed with another dynamical system that brings the robot towards regions of low uncertainty. From a control point of view, this results in adding another attractor field that is proportional to the gradient of the variance manifold \cite{Franzese2021ILoSA} according to
\begin{equation}
    \Delta \bm{x}_\mathrm{stable}(\bm{x})=
    -\alpha \nabla \Sigma %(\bm{x})\\
    =\alpha \left( 2 \bm{k}_{*}^\top \bm{K}^{-1}\frac{\partial\bm{k}_{*}}{\partial \bm{x}}\right),
\label{eq:variancegrad}
\end{equation}
where $\bm{x}$ is the evaluation point, and $\alpha$ is an automatically modulated constant which ensures that the product of $\Delta \bm{x}_s$ with the robot impedance $\bm{K_s}$ is never higher than a set threshold. 
This repulsive field can be seen as a \emph{behavioural} stiffness: considering a variance manifold as a potential energy, similar to elastic energy, the robot is always acting towards the minimization of this quantity; similarly, the lower level control, ``the muscles'', is trying to converge to the attractor in order to minimize its \emph{physical} tension.  
Thus, the Minimum Uncertainty Dynamical System (\emph{MUDS}) can be summarized as
\begin{equation}
    \bm{x}_\mathrm{des}=\bm{x}+\Delta\bm{x}(\bm{x})-\alpha \bm{\nabla} \Sigma\bm (\bm{x}).
\end{equation}
The superposition of this field is not conflicting because when close to the data, the prediction is non-zero while the uncertainty is zero with a small gradient. As the uncertainty increases, the prediction starts vanishing towards the mean of the independent Process (in our case zero) while the stabilization field increases its magnitude. This results in redirecting the robot towards regions of low uncertainty. 

\begin{algorithm}[t]
\SetAlgoLined
\DontPrintSemicolon
\textbf{a) Kinesthetic Demonstration(s)} \\
\While{Trajectory Recording}{ 
{Receive$\left(\bm{x}_t, \,\sin(\bm{\theta}^\mathrm{dem}(\bm{x}_t)), \, \cos(\bm{\theta}^\mathrm{dem}(\bm{x}_t)), \,w^\mathrm{dem}(\bm{x}_t)\right)$} \\
 $\Delta \bm{x}^\mathrm{demo}(\bm{x}_{t-1})=\bm{x}_t-\bm{x}_{t-1}$\\
 }
{Train(GPs)} \\
\textbf{b) Interactive Corrections} \\
\KwData{$\Delta \bm{x}^\mathrm{dem}$, $\gamma^\mathrm{dem}$, $\sin{\bm{\theta}^\mathrm{dem}}$, $\cos{\bm{\theta}^\mathrm{dem}}$, $w^\mathrm{dem}$} 
\While{Control Active}{
Receive($\bm{x}$) \\
\If{Received Human feedback $\Delta \bm{x}^c$, $\gamma^c$, $w^c$}{
    {Correct}($\Delta \bm{x}^c \xrightarrow{} \Delta \bm{x}^\mathrm{dem}$, $\gamma^c \xrightarrow{} \gamma^\mathrm{dem}$, $w^c \xrightarrow{} w^\mathrm{dem}$)\\
    % {Correct}($\gamma^c \xrightarrow{} \gamma^d$) \\
    % {Correct}($w^c \xrightarrow{} w^d$) 
    }
$[\Delta \bm{x}, \Sigma$] = GP$_{\Delta \bm{x}}(\bm{x})$ \\
$\gamma$= GP$_{\gamma}(\bm{x})$ \\
$w_\mathrm{des} =$ GP$^{MU}_w(\bm{x})$ \\
$[\sin(\bm{\theta}), \cos(\bm{\theta})]=$ GP$^{MU}_\theta(\bm{x})$\\
$\bm{\theta}_\mathrm{des}=$ $\operatorname{arctan2}\left(\sin(\bm{\theta}), \cos(\bm{\theta})\right)$\\
$\bm{x}_\mathrm{des}=\bm{x}+ \gamma \Delta \bm{x} -\alpha\nabla \Sigma$\\
{Send}($\bm{x}_\mathrm{des}, \bm{\theta}_\mathrm{des}, w_\mathrm{des}$)
}
 \caption{Teaching Framework}
 \label{algo::FG}
\end{algorithm}

\subsection{Minimum Uncertainty Inference}
\label{sec::mapping}
When learning a complex task like a fluent P\&P, the dynamics of the end-effector position have to be augmented with the dynamics of the gripper orientation and width. Because in a trajectory the dynamics of the orientation and gripper are coupled with the dynamics of the end-effector, we decided to learn the controlled action as a function of the robot's position with another GP. 
However, if the predictions are done based on the current position, when outside of the region of certainty, the robot would output the mean of an independent Process (i.e., zero radians for the orientation along all three axes and maximum gripper width) which could lead to an undesirable generalization, e.g., tilting or dropping objects. 
In order to solve this problem, we propose a \emph{minimum uncertainty inference}, 
obtained by projecting $\bm{x}$ in the highest correlated sample of the database according to
\begin{equation}
  \bm{x}= \underset{\bm{\xi}_i}{\operatorname{argmax}} \left(k\left(\bm{x},\bm{\xi}_i\right)\right)
\end{equation}
where $k$ is the 
kernel function with the optimized hyper-parameters.
This minimum uncertainty inference can be interpreted as a ``mental'' projection of the robot's current state on the highest correlated state (according to the kernel function) collected during the demonstration(s).  
The aim is to explicitly avoid extrapolating outside the original demonstrated data while still using the property of a smooth regressor of the GP. This behaviour also matches the philosophy of actively taking actions that would always minimize the uncertainty on the current robot state.  
When the evaluation of the GP is performed with this minimum uncertainty rule, we denote them with the superscript $MU$.

In order to fit the desired angles with a regressor, it is necessary to have a smooth and \emph{continuous} representation of the angles. To this end we fit both $\sin(\bm{\theta})$ and $\cos(\bm{\theta})$ transformations of the Euler angles and convert them back after the MU inference during robot control (l.~17 Alg.~\ref{algo::FG}).   
\label{oriAndgrip}

\subsection{Interactive Policy Correction with Human-in-the-Loop}
\label{sec::interactive}
After learning from kinesthetic demonstrations the desired transition $\Delta \bm{x}$, Euler angles $\bm{\theta}$ and the gripper width $w$ in the different points of the recorded trajectory, we still need to allow the user to correct the policy during the robot execution. 
% Considering that the goal is to speed up the motion beyond the capabilities of a person, it is not safe anymore to use a direct kinesthetic feedback for shaping the policy, like in \cite{kastritsi2018progressive}. 
Our goal is to obtain a fast continuous picking operation. With increasing velocities, kinesthetic interactions with a robot manipulator can become unsafe, and tuning both the attractor and gripper locally becomes very challenging. Furthermore, it also gives rise to ambiguity on the interpretation of the interaction forces as intended corrections or undesired disturbances \cite{kastritsi2018progressive}.
For this reason, we opted for teleoperated corrections  on the desired movement, local velocity and gripper width. %, similarly to what a teacher would do while correcting the movements of an apprentice. 
% Similarly when a teacher is correcting the movements of an apprentice, they are giving corrective feedback  
Thus, due to the necessity of modifying the magnitude of the attractor distance proportionally in \emph{all directions} (when higher/lower velocity are requested), a \emph{scaling} factor is learned as a function of the position, resulting in a desired attractor 
\begin{equation}
    \bm{x}_\mathrm{des}=\bm{x}+f(\bm{x})=\bm{x}+\gamma(\bm{x})\Delta\bm{x}(\bm{x})-\alpha \bm{\nabla} \Sigma\bm (\bm{x})
\end{equation}
where $\gamma(\bm{x})$ is the attractor scaling factor.
With this formulation, corrections can be allocated in the 3 different components of the vector or on the total magnitude of the vector itself. Overall, corrections are provided to the output values $\bm{y}_{demo}$ of the different GPs for the attractor distance $\Delta \bm{x}$, scaling factor $\gamma$ and the width of the gripper prongs $w$, all of which are initialised with the kinesthetic demonstration. With the evaluation of the kernel, the corrective input can be smoothly spread to surrounding data points in accordance with their correlation. 
The update rule was thus chosen as
\begin{equation}
\label{eq::update}
\bm{y}^\mathrm{demo}= \bm{y}^\mathrm{demo} + \bm{k}^{n}_{*}(\bm{\xi},\bm{x}) \epsilon_\mu % \epsilon_\mu,
\end{equation}
where $\bm{k}^{n}_{*}$ is the correlation vector $\bm{k}_{*}$ normalised such that $\sigma_f=1$, and $\epsilon_\mu$ is the given correction provided at $\bm{x}$. 

It has previously been shown that spreading the corrections on the database is more user-friendly, as well as time and data efficient \cite{Franzese2021ILoSA} than a simpler data aggregation \cite{kelly2019hg}, since otherwise the GP model would essentially average between the different outputs for a given input, leading to a slow learning.
Additionally, this constraint of spreading the corrections only on existing points of the database avoids to modify the shape of the variance manifold, keeping the motion always close to the kinesthetic demonstration, according to \eqref{eq:variancegrad}, while still shaping the motion dynamics, encoded in $\gamma(\bm{x})\Delta\bm{x}(\bm{x})$. 
\begin{figure}
    \centering
    \includegraphics[width=0.95\columnwidth]{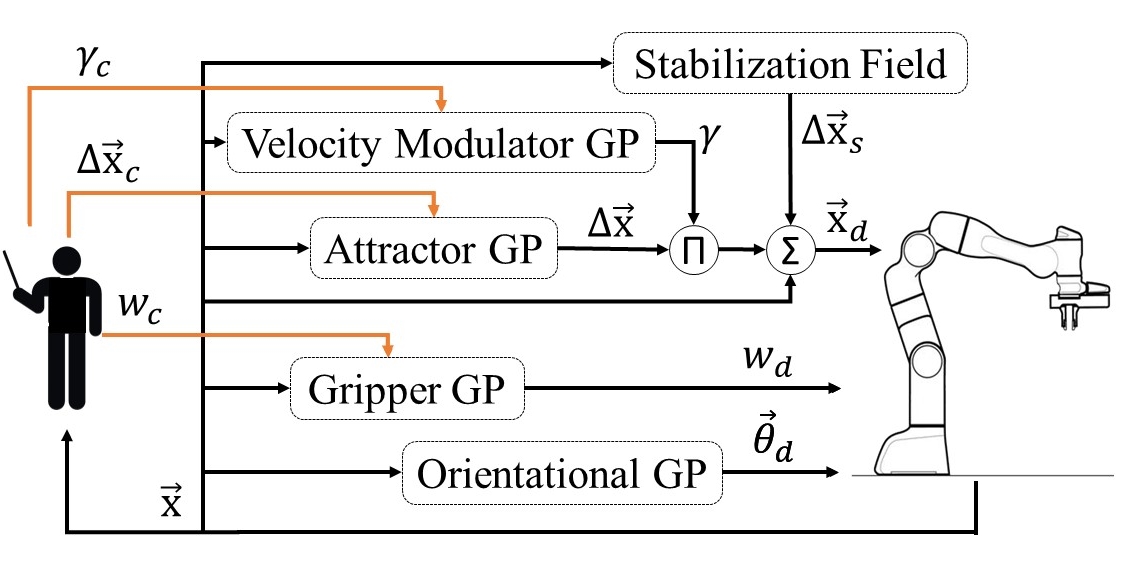}
    \caption{A schematic representation of the human-in-the-loop giving corrections to the learned policy. The human has a visual feedback of the current robot motion and gives corrections with a joystick. 
    }
    \label{fig:my_label}
\end{figure}

\label{method}
\section{Validation Experiments}
Experiments were carried out to evaluate the effectiveness, usability and robustness of the method. 
In Sec. \ref{sec:exp1}, the framework's base functionality of taking slow demonstrations and allowing the correction of the dynamics through corrective feedback is tested, along with an ablation study to verify the utility of uncertainty minimization. 
 In Sec. \ref{sec:exp2}, a baseline comparison to a method that also addresses the problem of interactive velocity modulation is performed.
Sec. \ref{sec:exp3} analyses how well a learned policy can accommodate changes in object properties such as size and weight. 
In Sec. \ref{sec:exp4}, a straightforward generalization w.r.t.\ different object locations is briefly analysed. 
Lastly, in Sec. \ref{sec:exp5} a user validation study was carried out with non-experts to establish the usability of the proposed method. A video of the %learning and execution of the tasks
experiments can be found attached to this paper \footnote{\url{https://youtu.be/XoW6AkK793g}}.

% \subsection{Robot Setup}
We used the 7 DoF Franka-Emika Panda with an impedance controller and a ROS communication network for the online attractor update with a frequency of \SI{100}{\Hz}. 
%Control of the robot's movement was carried out in real-time, whereas the control of the gripper could not be carried out in real-time due to the internal communication protocol. 
Furthermore, in order to avoid overloading the GP with superfluous data, the recording of the trajectory is carried out at \SI{10}{\Hz} considering that whatever the human is showing at higher frequency is noise that would anyway be filtered out by the GP fitting and the impedance policy.

A wireless Logitech F710 Gamepad was used for teleoperated corrections. %Our method, however, is agnostic to the input modality, since we only need the error between the current prediction and the desired output. 
The Gamepad was chosen due to the number of required inputs, it being an established ergonomic input device in the gaming industry, as well as ensuring that users remain at a safe distance from the robot at all times considering the high-speed motion dynamics. Due to the limited number of continuous inputs, both the gripper and scaling factor corrections are provided through discrete increments. 
The attractor corrections are provided through the continuous inputs of the two thumbsticks, with the movement in the $x$-$y$-plane regulated by the left thumbstick and the height regulated by the right thumbstick.  
As an added safety feature, one of the triggers was utilised as a \emph{safety button} which, when released, ends the execution of the algorithm, halting the robot. 
Lastly, users can comfortably start the execution from any point along the trajectory as well as bring the robot to the start of the trajectory. 
As a final remark, it is worth underlining that the capability of correcting the orientation after the demonstration was not enabled due to the limitations of the teleoperation interface, not due to any limitations surrounding the algorithm itself and is thus left to future work.

\subsection{Interactive Fluent Pick \& Place with MUDS}
\label{sec:exp1}

\begin{figure}[!t]
    \centering
    \includegraphics[width=0.8\columnwidth]{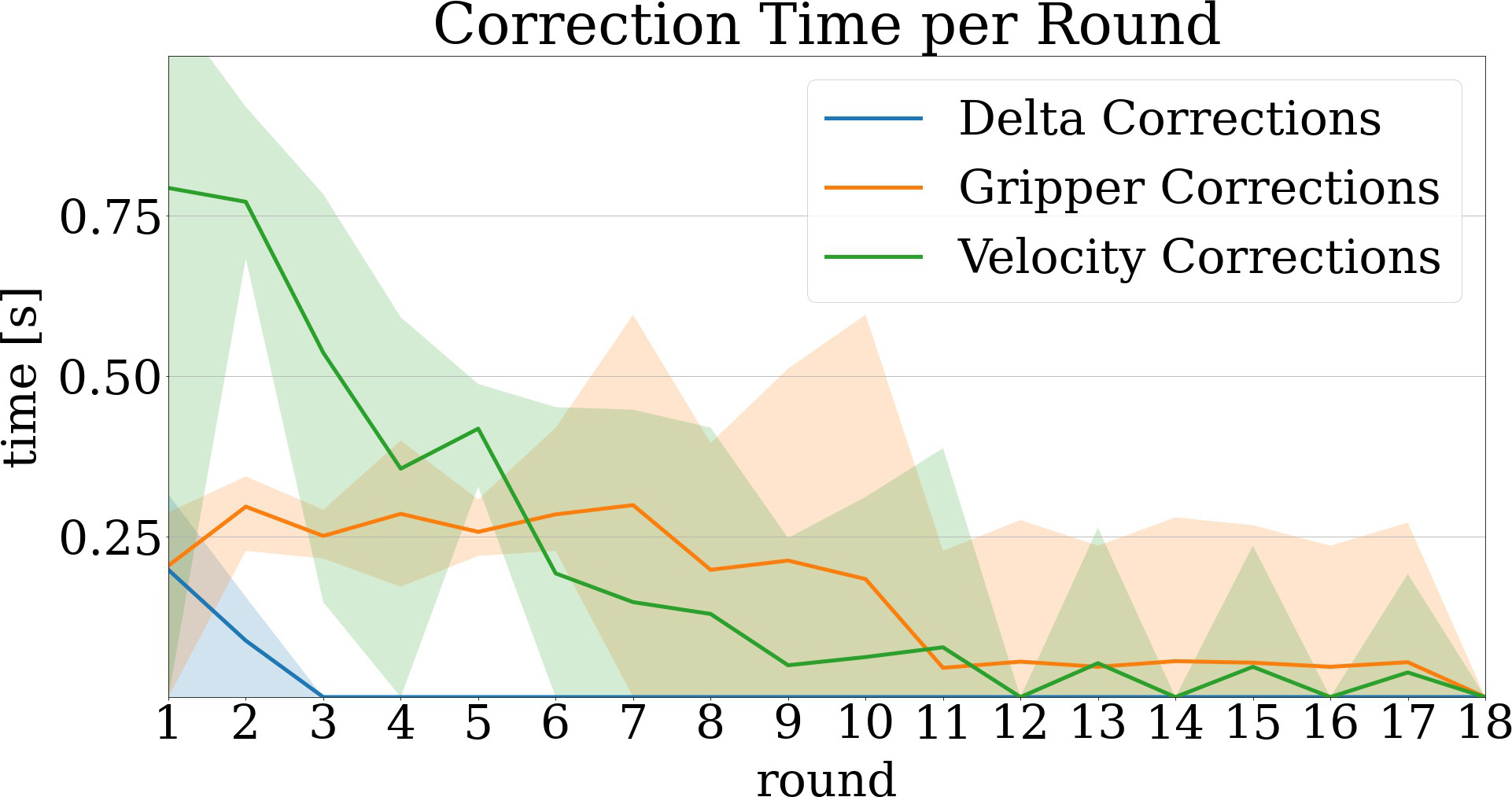}
    \caption{Range of correction times per round for each aspect depicted by the shaded areas, with the average times depicted by the solid lines. Statistics made over 5 repetitions.}
    \label{fig:correctionsPoC}
\end{figure}

\begin{figure*}[!t]
    \centering
    \includegraphics[width =0.85\linewidth]{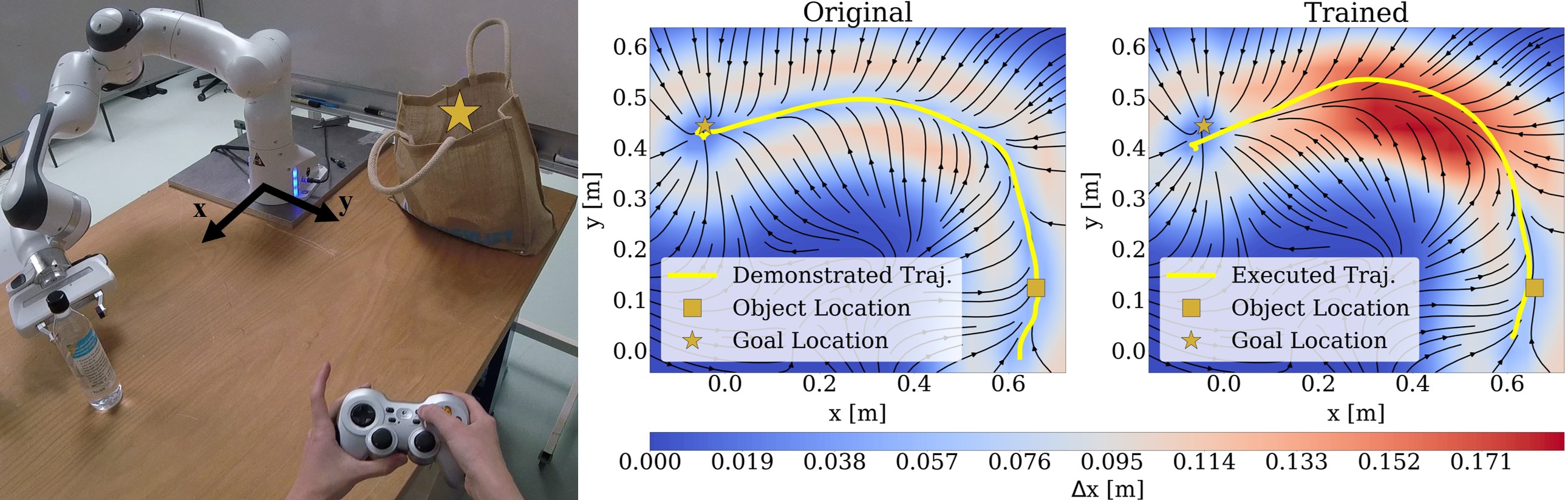} 
    \caption{Use case of robot assistance in grocery packing. In the attractor vector-field the arrows denotes the direction of the attractor and the color gradient denotes the magnitude of the attractor. The vector field based on original demonstration, with the demonstrated trajectory is compared with the one after training, with the executed trajectory.}
    \label{fig:vecFieldPoC}
\end{figure*}

For this experiment, a\emph{ single demonstration} was provided wherein the end-effector orientation, gripper width, and attractor distance are obtained and used for initialising the respective GP models. 
The goal of the task is to i) reduce the execution time by 4 times w.r.t.\ the demonstration time of the motion with kinesthetic teaching, and ii) have an execution time of \SI{3}{\second} or less. We repeated the experiment 5 times. 

Within less than \SI{3}{\minute} it was possible to fully train the robot to pick \& place the object with the desired performance, four out of five times.
Only a fraction of that time was needed for the demonstration (avg.\ \SI{11}{\s}) and explicit feedback from the human (avg.\ \SI{6.8}{\s}).
This points towards primarily needing fine-tuning corrections from the human, which is further supported by the time spent giving corrections for each of the three correctable aspects (see Fig. \ref{fig:correctionsPoC}).

It is worth noting that a correction round refers to an execution of a trajectory with optional user corrections, which can be stopped at any point of the execution and not just at the goal.
The time spent correcting the attractor was minimal, as it was only required around the moment when the object is reached.
This is because the human tends to stop at the object during the demonstration to avoid knocking it over and to deal with the closing of the gripper.
To avoid that the motion stops, minor corrections to the attractor were provided for ensuring it follows the desired continuous picking motion. 
Afterwards only corrections for the gripper and scaling factor are provided.
Whenever corrections to the scaling factor were provided, resulting in higher velocity, corrections to the gripper had to be provided as well to offset the communication delay of the gripper. 
%Once the desired velocity was achieved the final corrections were directed towards fine-tuning the gripper timing.
Due to the unreliability of the gripper, despite corrections to the timing, the gripper still sometimes closed at the incorrect moment.
Nevertheless, after corrections, an average success rate of $82\%$ out of 10 autonomous executions of 5 different trained policies (41 successes over 50 executions in total) could still be achieved. 
For the complete performance details, please refer to Tab.~\ref{tab:performancePoC}. 

To verify the existence of gripper unreliability we measured the delay between sending the command for closing the gripper and the actual moment of closing. Measurements were gathered from 20 rollouts. While the average delay was \SI{0.93}{\second}, it ranged from \SI{0.56}{\second} to \SI{1.54}{\second}. Considering this stochasticity, the best strategy is to push the object at non-zero velocity for a long enough time so that it encompasses the possible moments at which the gripper might close. 

\begin{table}[!t]
    \caption{Method Performance (5 demos, 50 executions)}
    \centering
    \begin{tabular}{|c|c|c|c|c|c|}
        \hline
         & \textbf{Demo \scriptsize[s]} & \textbf{Fdbk \scriptsize[s]} & \textbf{\makecell{Total\\ Time \scriptsize[s]}} &  \textbf{Rounds} & \textbf{\makecell{Success\\Rate \scriptsize[\%]}} \\
         \hline
        \textbf{Max} & 11.70 & 10.324 & 165.44 & 17.0 & 100 \\
        \hline
        \textbf{Mean} & 10.94 & 6.796 & 97.47 & 10.4 & 82 \\
        \hline
        \textbf{Min} & 10.10 & 4.560 & 66.61 & 6.0 & 50 \\
        \hline
    \end{tabular}
    \label{tab:performancePoC}
\end{table}
One of the main concerns when increasing the velocity along a trajectory is diverging from said trajectory, particularly in curves. While the shape of the trajectory did change slightly, divergence from the trajectory could be avoided thanks to the uncertainty minimisation even when the attractor magnitude was noticeably increased compared to the original demonstration. 
This can be observed within the attractor vector fields in Fig.~\ref{fig:vecFieldPoC}.
This is an important feature of the proposed method, opening an alternative to many methods that are not dealing with covariate shift when they try to generalize. The goal was to show that even if the dynamics of the trajectory are modified, the obtained trajectory is not changing much, resembling the original demonstration.  

To further evaluate the benefit of the uncertainty minimisation on the training as well as on the final execution, we performed an ablation study. 
The desired policy was trained once with the uncertainty minimisation active (w/ UM) and once without it (w/o UM). It was observed that the uncertainty minimisation made the training easier since it kept the robot close to the demonstrated trajectory. This translated to a shorter training time of \SI{70}{\second} w/ UM, whereas w/o UM \SI{218}{\second} were needed.
We then performed two tests for observing the effect on the execution; one with a perturbation to the robot's initial position and one without such perturbation. The policies were rolled out 20 times each. The effect of the uncertainty minimisation was observed in the success rates of the P\&P as well as the average distance error (ADE) of the executed trajectory w.r.t.\ the demonstrated trajectory. Without perturbations the policy w/ UM achieved the higher success rate of 95\% and lower ADE of \SI{0.023}{\metre} whereas the policy w/o UM only achieved a success rate of 45\% and an ADE of \SI{0.051}{\metre}. Similar results are observed when the perturbation is added, where the success rate of the policy w/ UM was 100\% and the ADE was \SI{0.034}{\metre} whereas the policy w/o UM only achieved 50\% and had an increased ADE of \SI{0.090}{\metre}.

For an evaluation of its benefit for reaching a goal while rejecting disturbances we would like to refer the reader to our prior work \cite{Franzese2021ILoSA}.

\subsection{Baseline Comparison}
\label{sec:exp2}
We compare to a state-of-the-art approach in interactive dynamics modulation presented in \cite{kastritsi2018progressive}. 
The base method was replicated based on the details given in the paper with the only major change being that we do not learn the orientation with the DMPs.
Since the focus of this baseline comparison was on the modulation of the translational dynamics, the gripper and orientation were controlled with the GPs, conditioned on the robot's current position for all the tests.
Corrections were given with the joystick in both cases out of safety concerns when the robot is moving at high velocity.

We initialized the DMP, a version of our algorithm using the scaling factor (\emph{V1}) and a version without the scaling factor (\emph{V2}) with a single demonstration of picking the object given along the $y$-axis. Then, the object was displaced \SI{7}{\cm} to the side ($x$-direction) to compare the ability of  both algorithms for reshaping and speeding up the motion. 

What could be noticed with the DMP-based approach is that when a correction in $x$-direction was given the robot would virtually stop and only occasionally move forward. The cause of this was determined to be the dot-product of the position error with the predicted velocity $\widetilde{\bm{p}}^\top\dot{\bm{p}}_d$. This is used for changing the temporal scaling factor $\tau$ of the DMPs, such that when the error is in the direction of the velocity the evolution of the DMP is sped up whereas in the opposite case, the evolution of the DMP is slowed down. In our case, although the predicted velocity along $x$ was very small, it was occasionally negative which could account for the undesired slowing down of the motion. Only in the moments when the velocity became positive along this axis did the robot move forward. As for speeding up, this was later possible along the $y$-axis, however, the generated acceleration was rather high even when a small correction was given. This is very likely due to the fast convergence of $\tau$. The total training time for a successful picking policy was \SI{197}{\second}. The final achieved execution time was \SI{8.43}{\second} which was $1.17$ times faster than the original.

With MUDS the correction in $x$-direction did not affect the motion in $y$-direction. Through the correction of the attractor distance along each of the axes, the shape of the trajectory along each of the axes could be easily altered.
When using the scaling factor $\gamma$ (\emph{V1}), the speed along each of the axes of motion increases proportionally.  
Alternatively, if one chooses to not use $\gamma$ and directly affect the velocity by changing the attractor distance along an axis (\emph{V2}), one can ensure that the corrections do not affect the remaining axes. Depending on whether the velocity increase should be proportional in all directions (e.g., speeding up a diagonal motion in $x$-$y$-direction) or only along a single axis, the two approaches of altering the velocity help account for both possibilities. With \emph{V1} \SI{48}{\second} were needed to train a successful picking policy whereas \SI{46}{\second} were needed for \emph{V2}. The final execution times were \SI{2.67}{\second} for \emph{V1} and \SI{2.24}{\second} for \emph{V2} which translated to an increase of speed by $3.71$ and $4.41$ times respectively.

\subsection{Interactive Adaptation to  New Object Properties}
\label{sec:exp3}
\begin{table}[!t]
    \centering
    \caption{Performance in Interactive Adaptation}
    \begin{tabular}{|c|c|c|c|c|}
        \hline
         & \textbf{\makecell{Rigid \\ (\SI{250}{\gram}) \\ \scriptsize %\textcolor{blue}{new} $|$ \textcolor{red}{adp}
         source
         }} 
         & \textbf{\makecell{Rigid \\ (\SI{900}{\gram}) \\ \, \scriptsize \textcolor{blue}{new} $|$ \textcolor{red}{adp}}} & \textbf{\makecell{Flexible \\ (\SI{100}{\gram}) \\ \scriptsize \textcolor{blue}{new} $|$ \textcolor{red}{adp}}} & \textbf{\makecell{Small \& \\def. (\SI{250}{\gram}) \\ \scriptsize \textcolor{blue}{new} $|$ \textcolor{red}{adp}}} \\
        \hline
        \textbf{\makecell{Correction \\ Time \scriptsize[s]}} & 
        51 
        & \textcolor{blue}{46} $|$ \textcolor{red}{0} & \textcolor{blue}{59} $|$ \textcolor{red}{38} & \textcolor{blue}{73} $|$ \textcolor{red}{24} \\
        \hline
        \textbf{Rounds}& 
        5 
        & \, \textcolor{blue}{5} $|$ \textcolor{red}{0} & \textcolor{blue}{7} $|$ \textcolor{red}{4} & \textcolor{blue}{8} $|$ \textcolor{red}{4} \\
        \hline
        \textbf{\makecell{Success \scriptsize[\%]}} &
        88 & \, \textcolor{blue}{96} $|$ \textcolor{red}{98} & \, \textcolor{blue}{98} $|$ \textcolor{red}{100} & \textcolor{blue}{98} $|$ \textcolor{red}{96} \\
        \hline
    \end{tabular}
    \label{tab:diffObj}
\end{table}

\begin{wrapfigure}{r}{0.18\textwidth}
    \centering
    \includegraphics[width=\linewidth, trim={1.4cm, 0.75cm, 1.4cm, 0.75cm}, clip]{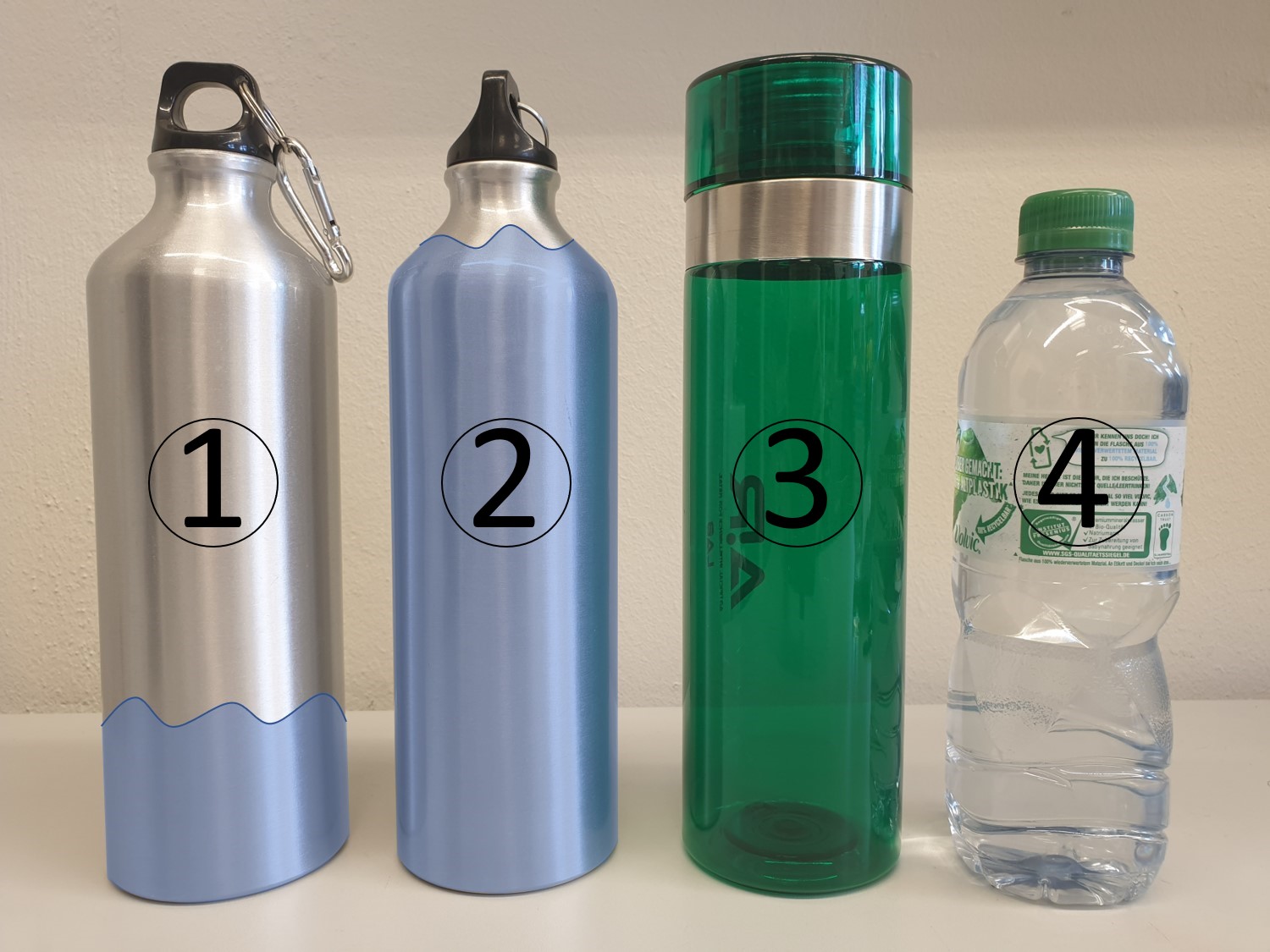}
    \caption{L-R: rigid (\SI{250}{\gram}), rigid (\SI{900}{\gram}), flexible (\SI{100}{\gram}), small \& deformable (\SI{250}{\gram})}
    \label{objects}
\end{wrapfigure}

It can be that we want to pick up a different object after having learned a desired P\&P behaviour. Even small changes in object properties can result in failure when using the same policy. Rather than demonstrating and retraining the strategy for every new object, or relying on hard-coded rules to adapt to these changes, corrections can be used to adapt the learned policy. A selection of four different objects was taken (seen in Fig.~\ref{objects}) to make a comparison of training from a new demonstration (\textcolor{blue}{new}) and training a policy by adapting an existing policy (\textcolor{red}{adp}), as reported in Table \ref{tab:diffObj}.

For the latter case, the initial policy was trained on a rigid water-bottle with a weight of \SI{250}{\gram} (\ding{172} in Fig.~\ref{objects}), our `source' object. Once a satisfactory policy was achieved, the training object was swapped out for another object. The policy was then executed and corrected if necessary. Corrections were provided until the policy was successfully executed with the new object, after which an evaluation of the performance was performed. Subsequently, a different object was swapped in and the learned policy was \emph{reset to the initial policy}.

For each new object, the policy could be successfully corrected.
For the same object but with a greater weight \ding{173} the initial policy carried out the policy successfully in the first execution, hence it was deemed that no corrections were necessary.
For the flexible object \ding{174} due to its lighter weight and ease at which it could be knocked over, minor corrections to both the velocity and gripper had to be given. 
Lastly, for the deformable object \ding{175} it was necessary to reduce the speed for a successful picking. Otherwise, the object kept being knocked over upon impact due to its smaller support polygon. Nevertheless, for all three objects with their different properties it was possible to alter the policy within less time than what is needed for training from a new demonstration (see Tab.~\ref{tab:diffObj}). 

It is important to note that the strategies for the separate objects are not stored as this would require a further form of knowledge representation or policy parametrization, which is outside the scope of this work. This evaluation does, however, show that adapting an existing policy is faster than learning from scratch, which can be beneficial for gathering knowledge more quickly.

\begin{table*}[!t]
    \caption{Performance of Non-Experts Who Successfully Finished the Task}
    \centering
    \begin{tabular}{|c|c|c|c|c|c|c|c|c|c|c|}
        \cline{2-9}
        \multicolumn{1}{c|}{}& \multicolumn{4}{c|}{T1: With Attractor Scaling } & \multicolumn{4}{c|}{T2: Without Attractor Scaling} \\
        \cline{2-9}
        \hline
         & \textbf{\makecell{Demo\\Time \scriptsize[s]}} & \textbf{\makecell{Training\\Time \scriptsize[s]}} & \textbf{Rounds} & \textbf{\makecell{Exec.\\Time \scriptsize[s]}} & \textbf{\makecell{Demo\\Time \scriptsize[s]}} & \textbf{\makecell{Training\\Time \scriptsize[s]}} & \textbf{Rounds} & \textbf{\makecell{Exec.\\Time \scriptsize[s]}} \\ 
         \hline
        \textbf{Max} & 34.10 & 600.00 & 36.00 & 4.97 & 14.90 & 285.00 & 23.00 & 4.00 \\
        \hline
        \textbf{Mean} & 13.04 & 323.30 & 19.40 & 3.42 & 8.63 & 121.22 & 9.11 & 2.81\\
        \hline
        \textbf{Min} & 6.40 & 129.00 & 6.00 &  2.17 & 3.90 & 0.00 & 0.00 & 2.07\\
        \hline
    \end{tabular}
    \label{tab:performanceNonExperts}
\end{table*}

\subsection{Generalizing to New Positions}
\label{sec:exp4}
An important point of any algorithm is the generalisation capability. The above experiments were confined to policies trained within a global frame, making their generalizability limited. This can be overcome by using the position w.r.t.\ a local reference frame as input. To this end, a minor alteration had to be made where two policies were learned; one w.r.t.\ a local frame within the target object, and one w.r.t.\ a local frame at the goal. To determine which policy should be used when, a simple heuristic was applied which stated that the robot first moves w.r.t.\ the object and after picking it up moves w.r.t.\ the goal \cite{franzese2020learning}. It is important to note that with the current approach there is a limit to how much the relative distance between the two frames can be changed w.r.t.\ the demonstrated one since when switching the frame of reference, the new policy must remain confident 
otherwise it will arrest the motion for safety.

To validate this extension we performed a short experiment where we trained the two policies (w.r.t.\ the object and w.r.t.\ the goal) with the frames fixed in one position. After the policy was successfully trained we placed the object in 20 different locations. The distance of these positions from the training location were taken from the ranges $x\in [-0.26; 0.02]$, $y\in [-0.30; 0.28]$, and $z\in [0; 0.08]$ all while considering locations physically feasible for the robot.

The total training time amounted to \SI{99.4}{\second} of which \SI{78.9}{\second} were needed for the corrections. Out of the 20 executions 13 were successful without any external influence, and 3 were successful once the human physically guided the robot into the region of certainty. For the latter 3, this was in fact a desired behaviour and a design choice to ensure that the robot does not generalise and potentially behave in an unsafe manner in situations it has never seen. If a person wants to add information on how to behave in these areas, this can be done by adding new points as was addressed in \cite{Franzese2021ILoSA}, but this was not the focus of the proposed method. 
The remaining 4 executions resulted in clear failure. Out of these, 2 were in the case where the object was placed at a greater height than the demonstration. After successfully picking up the object, the robot proceeded to get stuck against the surface of the table since the policy w.r.t.\ the goal dictated that it should be following a trajectory that was below its current position.

\subsection{Are Humans great teachers? A User Study}
\label{sec:exp5}
Since the aim of the proposed method is to enable people, who may not have a background in robotics and machine learning, to teach a robot, a preliminary user validation study was carried out. A total of ten participants aged 23 to 28 took part in this study (approved by TU Delft HREC). The same setup as in Fig.~\ref{fig:vecFieldPoC} was used, with the bag being replaced by a small square tower to provide a clearer goal.
Half an hour of familiarisation with the setup was given before the actual trials began. There were two trials of ten minutes which were presented in a \emph{randomised} order.
In one trial (\emph{T1}), users were required to perform a kinesthetic demonstration at a speed they were comfortable with.
Afterwards, they had the possibility to correct the demonstration with the possibility to scale the attractor distance. To ensure that the main contribution to the velocity resulted from the scaling factor, the attractor $\Delta \mathbf{x}$ itself was bounded to \SI{4}{\cm}.  
In the other trial (\emph{T2}), users were required to provide a \emph{fast} kinesthetic demonstration. The attractor for this trial was left unbounded and any corrections for the velocity had to be performed by directly altering the attractor in the three Cartesian directions. \emph{A trial was considered successful if the final trajectory execution time was \SI{4}{\second} or less}. The goal of this study was two-fold; i) verifying the feasibility of allowing non-experts to teach the robot non-zero-velocity P\&P and ii) determining which correction approach users may prefer.
In terms of performance, all participants were able to successfully pick \& place the object in T1. Only one was unable to reach the \SI{4}{\second} goal. For T2, only one was unable to teach the task successfully. 
 
Nevertheless, overall good teaching performance could be observed in both trials. For T1, users were able to teach the task within, on average, \emph{\SI{5.4}{\minute}} with 19 correction rounds. The average time at which the robot could successfully pick \& place the object that they could teach was \SI{3.4}{\second} with the \emph{best time being \SI{2.2}{\second}}. For reference, the time needed to demonstrate the behaviour at a fast pace in T2 was at best \SI{3.9}{\second}, but generally participants needed more than \SI{5}{\second} to carry out the demonstrations. \emph{It thus becomes clear that overall non-experts are not able to or are not comfortable with providing fast demonstrations}. 
Provided a faster demonstration, the time needed for corrections however did tend to be lower.

Participants were also asked which correction approach they preferred (T1 or T2). Within the group of participants, there was no clear preference towards one method or the other. There were, however, clear personal preferences.
Half preferred to correct the complete translational dynamics with one input, claiming that it made it easier for trajectory shaping or more intuitive for altering the velocity since it compared more closely to the controls that are familiar from video games. Meanwhile, the rest found it easier to focus on correcting one aspect at a time, thus preferring to first correct the trajectory before increasing the velocity with the scaling factor $\gamma$, since there was less chance of accidentally affecting the other aspect with the corrections. This means that by opting for only one correction approach, the performance and comfort of some people would be hampered. For this reason it is important that the method gives people the possibility of using either of the two approaches.

\section{Conclusions and Future Work}
\label{conclusion}
We demonstrated that the motion dynamics of a user's demonstration can be successfully altered in a non-uniform manner using teleoperated user corrections. This allows users to overcome the limitations they had during the demonstration and teach the actual desired behaviour. 
It further allows users to compensate for delays within the system which are not directly known to them but are observable in the system's performance.
Additionally, generalization to different object positions was obtained by switching between the two dynamical systems, learned in the respective reference frames. This proved how the variance minimization can be successfully used also to transition between two different frames. This opens many possibility of creating a sequence of multiple simpler dynamical systems for accomplishing complex robot tasks, i.e., assembling multiple components.

It was additionally shown that non-experts are able to successfully teach a non-zero-velocity motion for picking \& placing objects. Irrespective of their prior experience or lack thereof with robots, they were able to successfully train this complex task, teaching and correcting the motion dynamics of many degrees of freedom. 
It could be seen that when only using the kinesthetic demonstration, people generally could not attain the desired execution time even with a fast demonstration. However, with the help of corrections to the motion dynamics, an execution speed outside of their demonstration capabilities became achievable. 
Since people have different preferences of teaching and correcting robots, 
we concluded that the final framework requires the velocity corrections to be provided both in a coupled (with only $\Delta x$) and decoupled manner (with $\gamma$ and bounded $\Delta x$). 

Certain aspects remain to be addressed for better generalization and performance of the proposed framework. 
A next step would be to study how to obtain haptic corrections of the policy while ensuring a fast but safe human-robot interaction.
Further work is also needed in order to account for obstacles and reshape the vector field accordingly.

\bibliographystyle{IEEEtran}
\bibliography{biblio}

\begin{thebibliography}{10}
\providecommand{\url}[1]{#1}
\csname url@samestyle\endcsname
\providecommand{\newblock}{\relax}
\providecommand{\bibinfo}[2]{#2}
\providecommand{\BIBentrySTDinterwordspacing}{\spaceskip=0pt\relax}
\providecommand{\BIBentryALTinterwordstretchfactor}{4}
\providecommand{\BIBentryALTinterwordspacing}{\spaceskip=\fontdimen2\font plus
\BIBentryALTinterwordstretchfactor\fontdimen3\font minus
  \fontdimen4\font\relax}
\providecommand{\BIBforeignlanguage}[2]{{%
\expandafter\ifx\csname l@#1\endcsname\relax
\typeout{** WARNING: IEEEtran.bst: No hyphenation pattern has been}%
\typeout{** loaded for the language `#1'. Using the pattern for}%
\typeout{** the default language instead.}%
\else
\language=\csname l@#1\endcsname
\fi
#2}}
\providecommand{\BIBdecl}{\relax}
\BIBdecl

\bibitem{van2021robot}
J.~J. van Steen, N.~van~de Wouw, and A.~Saccon, ``Robot control for
  simultaneous impact tasks via {QP} based reference spreading,'' \emph{arXiv
  preprint arXiv:2111.05211}, 2021.

\bibitem{billard2019trends}
A.~Billard and D.~Kragic, ``Trends and challenges in robot manipulation,''
  \emph{Science}, vol. 364, no. 6446, p. eaat8414, 2019.

\bibitem{ravichandar2020recent}
H.~Ravichandar, A.~S. Polydoros, S.~Chernova, and A.~Billard, ``Recent advances
  in robot learning from demonstration,'' \emph{Annu. Rev. Control Robot.
  Auton. Syst.}, vol.~3, pp. 297--330, 2020.

\bibitem{Franzese2021ILoSA}
G.~Franzese, A.~M{\' e}sz{\' a}ros, L.~Peternel, and J.~Kober, ``{ILoSA}:
  Interactive learning of stiffness and attractors,'' in \emph{IEEE Int. Conf.
  Intell. Robots Syst. (IROS)}, 2021.

\bibitem{nemec2016speed}
R.~Vuga, B.~Nemec, and A.~Ude, ``Speed adaptation for self-improvement of
  skills learned from user demonstrations,'' \emph{Robotica}, vol.~34, no.~12,
  pp. 2806--2822, 2016.

\bibitem{kim2014catching}
S.~Kim, A.~Shukla, and A.~Billard, ``Catching objects in flight,'' \emph{IEEE
  Trans. Robot.}, vol.~30, no.~5, pp. 1049--1065, 2014.

\bibitem{salehian2016dynamical}
S.~S.~M. Salehian, M.~Khoramshahi, and A.~Billard, ``A dynamical system
  approach for softly catching a flying object: Theory and experiment,''
  \emph{IEEE Trans. Robot.}, vol.~32, no.~2, pp. 462--471, 2016.

\bibitem{bogdanovic2020learning}
M.~Bogdanovic, M.~Khadiv, and L.~Righetti, ``Learning variable impedance
  control for contact sensitive tasks,'' \emph{IEEE Robot. Autom. Lett.},
  vol.~5, no.~4, pp. 6129--6136, 2020.

\bibitem{haddadin2009requirements}
S.~Haddadin, A.~Albu-Sch{\"a}ffer, and G.~Hirzinger, ``Requirements for safe
  robots: Measurements, analysis and new insights,'' \emph{Int. Journal of
  Robotics Research}, vol.~28, no. 11-12, pp. 1507--1527, 2009.

\bibitem{koert2019learning}
D.~Koert, J.~Pajarinen, A.~Schotschneider, S.~Trick, C.~Rothkopf, and
  J.~Peters, ``Learning intention aware online adaptation of movement
  primitives,'' \emph{IEEE Robot. Autom. Lett.}, vol.~4, no.~4, pp. 3719--3726,
  2019.

\bibitem{nemec2018human}
B.~Nemec, N.~Likar, A.~Gams, and A.~Ude, ``Human robot cooperation with
  compliance adaptation along the motion trajectory,'' \emph{Autonomous
  Robots}, vol.~42, no.~5, pp. 1023--1035, 2018.

\bibitem{kastritsi2018progressive}
T.~Kastritsi, F.~Dimeas, and Z.~Doulgeri, ``Progressive automation with {DMP}
  synchronization and variable stiffness control,'' \emph{IEEE Robot. Autom.
  Lett.}, vol.~3, no.~4, pp. 3789--3796, 2018.

\bibitem{perez2020interactive}
R.~Perez-Dattari, C.~Celemin, G.~Franzese, J.~Ruiz-del Solar, and J.~Kober,
  ``Interactive learning of temporal features for control: Shaping policies and
  state representations from human feedback,'' \emph{IEEE Robot. Autom. Mag.},
  vol.~27, no.~2, pp. 46--54, 2020.

\bibitem{kronander2015incremental}
K.~Kronander, M.~Khansari, and A.~Billard, ``Incremental motion learning with
  locally modulated dynamical systems,'' \emph{Robotics and Autonomous
  Systems}, vol.~70, pp. 52--62, 2015.

\bibitem{Rasmussen2005}
C.~E. Rasmussen and C.~K.~I. Williams, \emph{Gaussian Processes for Machine
  Learning}.\hskip 1em plus 0.5em minus 0.4em\relax The MIT Press, 2006.

\bibitem{kelly2019hg}
M.~Kelly, C.~Sidrane, K.~Driggs-Campbell, and M.~J. Kochenderfer,
  ``{HG-DAgger}: Interactive imitation learning with human experts,'' in
  \emph{IEEE Int. Conf. Robot. Autom. (ICRA)}, 2019.

\bibitem{franzese2020learning}
G.~Franzese, C.~Celemin, and J.~Kober, ``Learning interactively to resolve
  ambiguity in reference frame selection,'' in \emph{Conf. Robot Learning
  (CoRL)}, 2020.

\end{thebibliography}
\end{document}